\documentclass{bmvc2k}

\usepackage{times}
\usepackage{epsfig}
\usepackage{graphicx}
\usepackage{amsmath}
\usepackage{amssymb}
\usepackage{booktabs}
\usepackage{subfigure}

\title{Hashmod: A Hashing Method for Scalable 3D Object Detection }

\addauthor{Wadim Kehl}{kehl@in.tum.de}{ 1}
\addauthor{Federico Tombari}{federico.tombari@unibo.it}{ 12}
\addauthor{Nassir Navab}{navab.cs.tum.edu}{ 1}
\addauthor{Slobodan Ilic}{slobodan.ilic@siemens.com}{ 3}
\addauthor{Vincent Lepetit}{lepetit@icg.tugraz.at}{ 4}

\addinstitution{
 Computer-Aided Medical Procedures, \\
 TU Munich, Germany
}

\addinstitution{
	Computer Vision Lab (DISI), \\ 
	University of Bologna, Italy
}

\addinstitution{
 Siemens AG \\
 Research \& Technology Center \\
 Munich, Germany
}

\addinstitution{
	Institute for Computer Graphics and Vision, \\ TU Graz, Austria
}

\runninghead{Kehl et al.}{Hashmod: Scalable 3D Object Detection}

\DeclareMathOperator*{\argmin}{arg\,min}

\begin{document}

\maketitle

\begin{abstract}
We  present a  scalable method  for detecting  objects and  estimating their  3D
poses in RGB-D data. To this  end, we rely on  an efficient representation of object  views 
 and  employ hashing techniques  to match  these views against  the input
frame  in  a scalable  way.  While  a similar  approach  already  exists for  2D
detection, we  show how to  extend it  to estimate the  3D pose of  the detected
objects.  In  particular, we explore  different hashing strategies  and identify
the one  which is  more suitable to  our problem. We  show empirically  that the
complexity of our method  is sublinear with the number of  objects and we enable
detection  and pose  estimation  of many  3D objects  with  high accuracy  while
outperforming the state-of-the-art in terms of runtime.
\end{abstract}

\section{Introduction}

Scalable 3D object detection and pose  estimation remains a hard problem to this
day. The recent  advent of low-cost RGB-D sensors boosted  the research activity
on  object instance  detection and  3D  pose estimation  even further,  allowing
state-of-the-art methods to robustly detect  multiple objects and estimate their
3D poses even under high levels of occlusion.  However, while  image recognition and
2D object recognition methods can now scale to billions of images or millions of
objects~\cite{Nister2006,Jegou2011,Dean2013,Norouzi2014},  3D  object  detection
techniques are still typically limited to ten or so objects.

Some  attempts  to  make  3D  object  detection  scalable  are  based  on  local
descriptions of 2D or  3D keypoints, since such descriptors can  be matched in a
sublinear  manner via  fast indexing  schemes~\cite{Muja2014}.  However,  computing such descriptors is expensive~\cite{Aldoma2013,Xie2013},   and  more
importantly,  they tend  to  perform poorly  on  objects without  discriminative
geometric or  textural features.   \cite{Brachmann2014,Tejani2014, Bonde2014} also  rely on
recognition of  densely sampled patches but are likely to work only when depth data
is available.   \cite{Lai2011} uses  a tree for  object recognition,  but still
scales  linearly  in  the  number  of objects,  categories,  and  poses.   Other
approaches  rely  on part-based  models~\cite{Savarese2007,Payet2011,Pepik2012},
which are  designed for category  recognition rather than  instance recognition,
and little concern is given to the  complexity, which is typically linear in the
number of objects.

Our approach  to 3D  object detection  is based  on 2D  view-specific templates
which    cover    the    appearance     of    the    objects    over    multiple
viewpoints~\cite{Hoiem2011,Nayar1996,Gu2010,Hinterstoisser2012,Aubry2014}.
Since viewpoints include the whole object appearance rather than just parts
of it, they can generally handle  objects with poor visual features, however they  have  not  been  shown  to scale  well  with  the  number  of  images so far. ~\cite{Tombari2013}.

We are  strongly influenced by \cite{Dean2013}  which showed impressive
results in terms  of 2D object detection by replacing  convolutions of templates
with  constant-time  hash  table  probes,  parsing  input  images  with  100,000
templates in  about twenty seconds  with respectable precision rates.   We apply
hash  functions~\cite{Gionis1999} to  image descriptors  computed over  bounding
boxes centered at each image location of the scene, so to match them efficiently
against a large descriptor database of model views.  In our work, we rely on the
LineMOD descriptor~\cite{Hinterstoisser2012},  since it  has been shown  to work
well   for   3D  object   detection,   although   other  descriptors   such   as
HOG~\cite{Dalal2005} could be certainly used as well.

\begin{figure}
	\begin{center}
		\includegraphics[width=5.5cm]{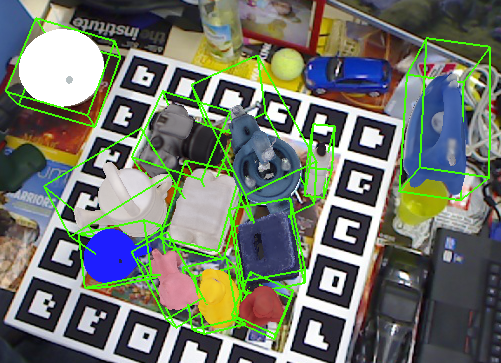} \hspace{0.5cm}
		\includegraphics[width=5.5cm]{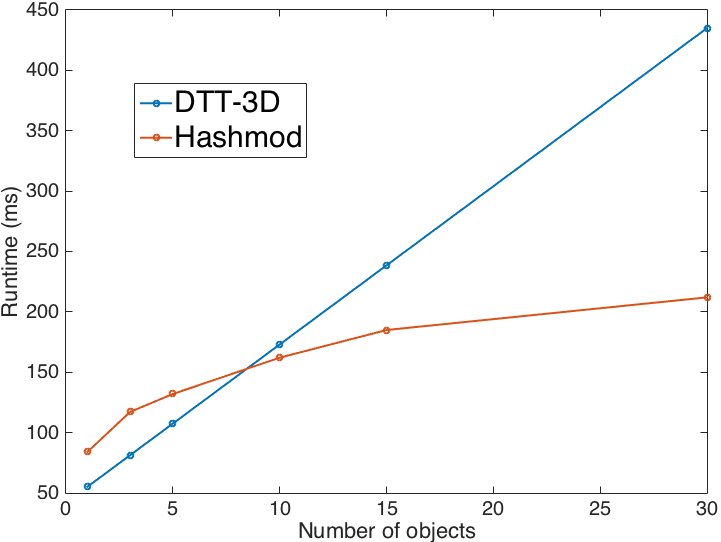}
	\end{center}
	\caption{Left: One frame of the ACCV12 dataset~\cite{Hinterstoisser2012} augmented with our detections. Right: Average performance of our approach with a given amount of objects in the database. We clearly scale sublinearly and outperform the state-of-the-art~\cite{Rios-Cabrera2013} with more than 8 objects, enabling detection of many 3D objects at interactive runtimes.}
	\label{fig:teaser}
\end{figure}

Our contribution is to present an efficient way of hashing such a descriptor: to
this end, we  explore different learning strategies to identify  the one that is
most suited  to our problem.   As shown in Figure~\ref{fig:teaser},  we outperform
the state-of-the-art template  matching method DTT-3D~\cite{Rios-Cabrera2013} by
intelligently hashing our descriptors, achieving sublinear scalability.

\section{Related work}

3D  object   detection  has  a  long   history.   Early  works  were   based  on
edges~\cite{Lowe91,Harris92},  then keypoint-based  methods were  shown to  work
reliably            when           distinctive            features           are
available~\cite{Nister2006,Wagner08,Tombari2010,Pauwels2013} and  robust schemes
for       correspondence       filtering        and       verification       are
used~\cite{Hao2013,Aldoma2013,Buch2014}.   Furthermore, they  are also  scalable
since they  can be reduced to  searching nearest neighbors efficiently  in their
feature  spaces~\cite{Muja2014,Jegou2011}.    However,  if  such   features  are
missing,  which is  actually  the case  for many  daily  objects, this  approach
becomes unreliable.

Template-based          approaches         then          became         popular.
LineMOD~\cite{Hinterstoisser2012} achieved  robust 3D object detection  and pose
estimation  by  efficiently  matching  templated  views  with  quantized  object
contours  and  normal  orientations.   In  \cite{Rios-Cabrera2013}  the  authors
further optimize the matching via cascades and fine-tuned templates to achieve a
notable run-time  increase by a  factor of  10.  Nonetheless, these  works still
suffer         from         their        linear         time         complexity.
\cite{Malisiewicz2011,Gu2010,Rios-Cabrera2014} show how  to build discriminative
models based on these representations using  SVM or boosting applied to training
data.  While  \cite{Malisiewicz2011,Rios-Cabrera2014} do  not consider  the pose
estimation  problem,  \cite{Gu2010}  focuses  on   this  problem  only  with  a
discriminatively trained mixture of HOG templates.  Exemplars were also recently
used for  3D object detection  and pose estimation in~\cite{Aubry2014},  but the
proposed approach still does not scale.

\cite{Brachmann2014,Tejani2014} use forest-based voting schemes on local patches
to detect and estimate 3D poses.   While the former regresses object coordinates
and conducts  a subsequent  energy-based pose estimation,  the latter  bases its
voting on  a scale-invariant  LineMOD-inspired patch representation  and returns
location and pose simultaneously.    \cite{Bonde2014} also uses Random Forests to
infer object  and pose,  but  via  a sliding window  through a  depth volume.
These methods remain slow, and it is not clear how they scale in performance
with the number of objects.

Over  the  last  years,  hashing-based   techniques  became  quite  popular  for
large-scale image  classification since they  allow for immediate  indexing into
huge  datasets.  Apart  from many  works that  focused on  improving hashing  of
real-valued  features into  more  compact binary  codes~\cite{Lin2014,Gong2013},
there has  been ongoing  research on  applying hashing  in a sliding
window scenario for  2D object detection: \cite{Dean2013} applies  hashing on HOG
descriptors computed from  Deformable Part Models to scale to  100,000 2D object
classes.   \cite{Aytar2014}  presents a  scalable  object  category detector  by
representing  HOG  sparsely  with  a  set of  patches  which  can  be  retrieved
immediately.

Hashing  has also  been used  for 3D  object detection:  \cite{Damen2012} hashes
paths  over edgelets in color images and allows for  real-time  3D object  detection,  however the  output remains in terms  of 2D locations only.  \cite{Cai2013}  applies uniform quantization to edge-based descriptors to immediately look up approximate nearest neighbors. Here we instead prefer hashing techniques in conjunction with a learning scheme since it tends to provide better accuracy by taking the actual data variety into account.

\newcommand{\Descs}{\mathcal{D}}
\newcommand{\bx}{\mathbf{x}}
\newcommand{\by}{\mathbf{y}}
\newcommand{\IB}{\mathbb{B}}

\section{Hashing for Object Recognition and 3D Pose Estimation}

\begin{figure}
	\begin{center}
		\includegraphics[height=2.5cm]{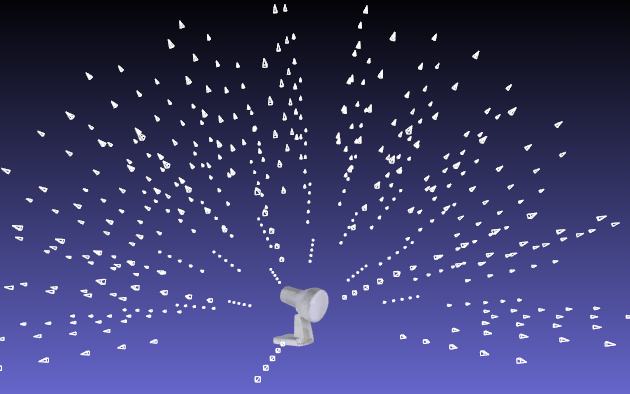} \hspace{0.5cm}
		\includegraphics[height=2.5cm]{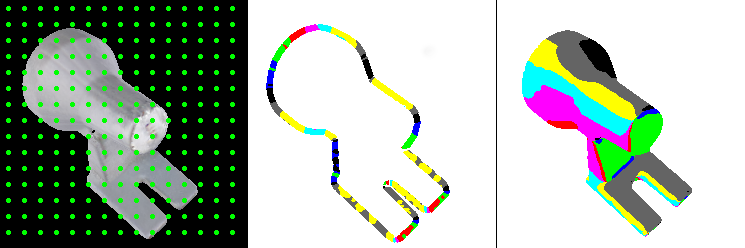}
	\end{center}
	\caption{Left: As in ~\cite{Hinterstoisser2012}, we compute LineMOD descriptors
		for synthetically rendered views sampled on hemispheres of several radii.  Right: One such synthetic  view of
		the `lamp' object  overlaid with the fixed grid for matching and the color-coded quantized orientations of image gradients and 3D normals used to compute the descriptor. }
	\label{fig:linemod}
\end{figure}

\begin{figure*}[t]
	\begin{center}
		
		\includegraphics[width=13cm]{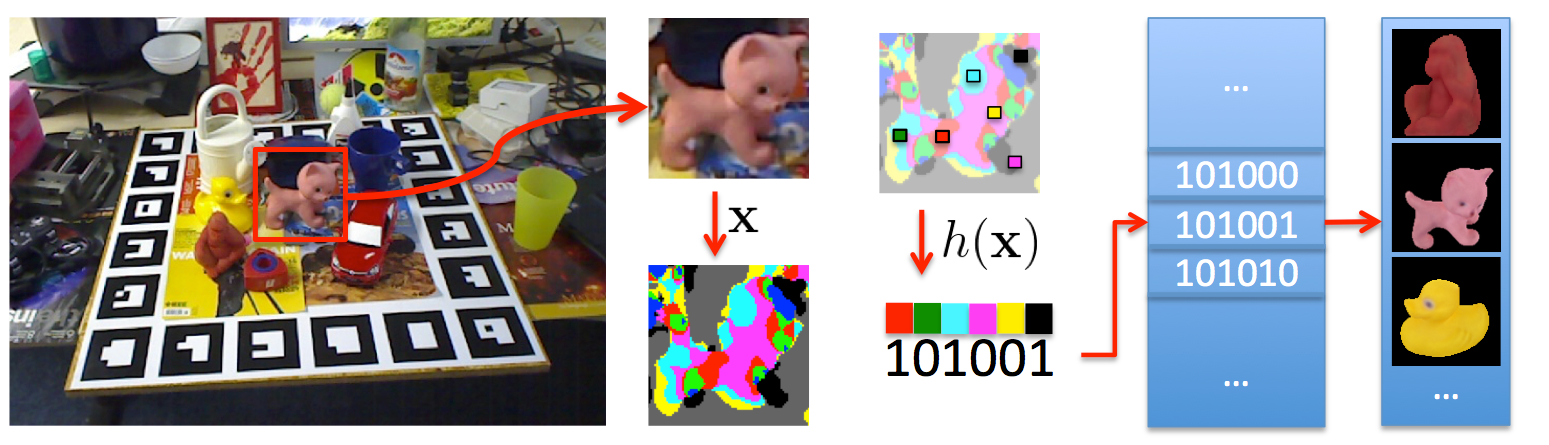}
	\end{center}
	\caption{Visualization of the hashing pipeline with one hash function $h$ and a
		key length of  $b=6$. At each sliding  window's position we extract  a LineMOD
		descriptor $\bx$  and sample  certain orientations  at specific  positions to
		form  a short  binary  string $h(\bx)$.  This  serves as  an  index into  the
		pre-filled hash table to retrieve  candidate views for further matching. Both
		the sample positions and orientations for $h$ are learned.}
	\label{fig:pipeline}
\end{figure*}

Given a database of $M$ objects, we synthetically create $N$ views for
each object from poses regularly sampled on a hemisphere of a given
radius, as shown in Figure~\ref{fig:linemod}. From this, we compute  a set  $\Descs$  of $d$-dimensional  binary
descriptors:
\begin{equation}
\Descs  = \left\{\bx_{1,1},...,\bx_{M,N} \right\} \;\;,
\end{equation}
where $\bx_{i,j} \in  \IB^d$ is the descriptor for the  $i$-th object seen under
the $j$-th  pose.  As already mentioned,  we use LineMOD in  practice to compute
these descriptors.  The LineMOD descriptor is a vector of integers between 0 and
16 and for each pixel it is either set to 0 when there
is no significant image gradient or depth data present or otherwise set to a value to represent a quantized orientation of either an image gradient (1-8) or 3D normal (9-16). We  concatenate the  binary representation  of these
integer values  to obtain the  binary strings  $\bx_{i,j}$. In the  remainder of
this work we will use the terms views, templates, and descriptors as synonyms.

Figure~\ref{fig:pipeline} gives an overview of our pipeline.  As usually done in
template-based approaches, we parse the image  with a sliding window looking for
the objects  of interest.  We extract  at each image location  the corresponding
descriptor  $\bx$.  If  the  distance  between $\bx$  and  its nearest  neighbor
$\bx_{i,j}$  in $\Descs$  is small  enough,  it is  very likely  that the  image
location contains object $i$ under pose  $j$.  As discussed in the introduction,
we want  to perform  this (approximate) nearest  neighbor search  by hashing  the descriptors.
Therefore,   we   explore  different   strategies   for   building  the   hashing
functions. This is done in the offline stage described below.

Also note that we tackle  the  issue  of object  scale  and  views of different
2D sizes
by dividing the  views up  into clusters
$\mathcal{D}_s  \subset  \Descs$  of  similar  scale $s$.   This  leads  to  $s$
differently-sized sliding windows during testing which extract differently-sized
descriptors on  which to  perform the hashing.  Moreover, to  increase detection
rates, we  assign a pre-defined  number of hash  functions per window  such that
they relate to random but overlapping subsets of $\mathcal{D}_s$. During testing,
we evaluate all  sliding windows with their associated  hashing functions, union
their retrieved views  and conduct subsequent matching. Lastly,  we determined a
good compromise for the key lengths by setting for each scale $s$ the key length
$b_s := \lfloor \log_2(|\mathcal{D}_s|) \rfloor$.

\subsection{Selecting the Hashing Keys}

During    our   offline    stage,   we    learn   several    hashing   functions
$h$~\cite{Gionis1999}.  As  shown in  Figure~\ref{fig:pipeline}, the  purpose of
each function is to immediately index  into a subset, often called a ``bucket'',
of $\Descs$ when applied to a  descriptor $\bx \in \IB^d$ during testing.  These
buckets are  filled with descriptors from  $\Descs$ with the same  hash value so
that we can restrict our search for  the nearest neighbor of $\bx$ to the bucket
retrieved via  the hashing function  instead of  going through the  complete set
$\Descs$.  It is  very likely, but not guaranteed, that  the nearest neighbor is in
at least one of the buckets  returned by the  hashing functions. 

In practice, a careful selection of  the hashing functions is important for good
performance.  Since the descriptors $\bx$  are already binary strings, we design
our hashing functions $h(\bx)$ to return a short binary string made of $b$ bits
directly extracted from $\bx$. This is a very efficient way of hashing and we will refer to these short strings as hash keys.

There is a typical trade-off between accuracy and speed: we want to retrieve only a handful of descriptors at each image location and the number of retrieved elements is governed by the hash key length and the distribution of the descriptors among the buckets. Since we use $b$ bits for the hash table, we span a table with $2^b$ buckets. If a key is too short, the number of buckets is too small and we store overproportionally many descriptors per bucket, increasing subsequent matching time after retrieval. If the key is too large, we might be more prone to noise in the bitstrings which may lead to wrong buckets, rendering false negatives more probable during testing. We thus want to select these $b$ bits in a  way such that we maximize the odds of  finding the  nearest neighbors of the input descriptors in the buckets while keeping the total amount of retrieved views to a minimum.

An exhaustive evaluation  of all the possible bit  selections to build
the  hash  keys  is  clearly intractable.  We  experimented  with  the
following variants:

\paragraph{Randomness-based selection (RBS)}
Given a set of descriptors, we  select the $b$ bits
randomly among all possible $d$ bits. As we will see later on, this selection strategy yields bad results since some bits are more discriminant than
others  in  our template  representation. Nonetheless, it  provides us with  a weak  baseline we  can compare
to and it outlines the importance  of a  more sophisticated approach  towards hash
learning.

\newcommand{\bit}{B}
\newcommand{\zero}{{\tt 0}}
\newcommand{\one}{{\tt 1}}

\paragraph{Probability-based selection (PBS)}
For this strategy, we  focus on the bits for which the probabilities of being $0$
and $1$ are close to $0.5$ with a given set of descriptors. We therefore rank
each bit $B$ according to its entropy $E = p(\bit = \zero) \ln p(\bit = \zero) +
p(\bit =\one) \ln p(\bit = \one)$ and take the best $b$. This strategy provides
a high accuracy since it focuses on the most discriminant bits. However, later evaluation will reveal that this strategy results in a high variance in the number of elements per bucket, rendering PBS inefficient in terms of runtime.

\paragraph{Tree-based selection (TBS)} 
This strategy is  inspired by greedy tree growing for  Randomized Forests. Starting with  a set of descriptors  at the
root, we  determine the bit that  splits this set  into two subsets with sizes as
equal as possible, and use it as the  first bit of the key.  For the second bit,
we  decide  for  the  one  that  splits those  two  subsets  further  into  four
equally-sized subsets and so  forth.  We stop if $b$ bits  have been selected or
one subset becomes empty.  This procedure alone  yields a balanced tree with leafs of
similar numbers of  elements. Each hash key can be regarded as a path down the tree and each leaf represents a bucket. Note that such a balanced  repartition  ensures  retrieval  and matching  at  a  constant  speed.  Formally, the $j$-th bit $B$ of the key is selected by solving:
\begin{equation}
\argmin_B \sum_i  \Big| |\mathcal{S}_L^B(N_i) | - |\mathcal{S}_R^B(N_i)|  \Big|  \;\; , 
\end{equation}
where $N_i \subset  \Descs$ is the set of descriptors contained by  the $i$-th node at
level  $j$, and  $\mathcal{S}_{\{L,R\}}^B(N_i)$  are the  two  subsets of  $N_i$ that go into the left and right child induced by splitting with $B$.

\begin{figure}[t]
	\begin{center}
		
		\includegraphics[width=8cm]{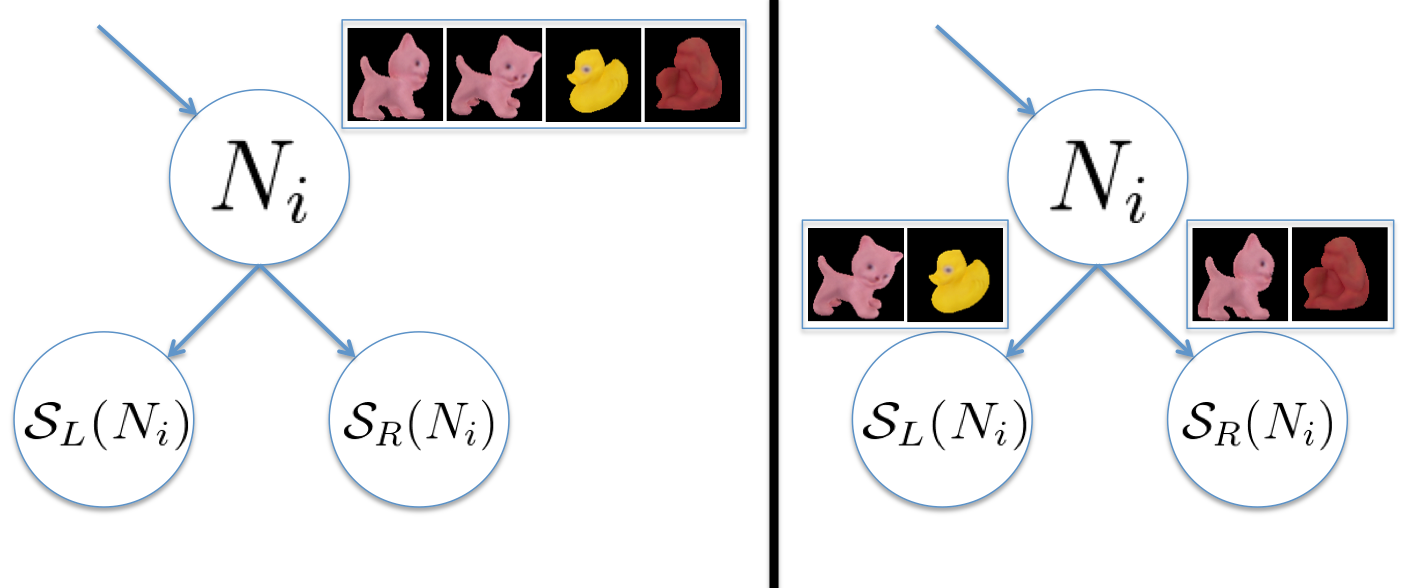}
	\end{center}
	\caption{The TBV strategy encourages descriptors for similar views to fall
		into different buckets.  This increases the chances to find a close
		descriptor when parsing the buckets.}
	\label{fig:tree}
\end{figure}

\paragraph{Tree-based selection with view scattering (TBV)}
We now further adapt the TBS strategy to our problem:  as
illustrated in Figure~\ref{fig:tree}, to improve detection rates we  favor similar views  of the
same  object to  go into different branches. The idea behind this strategy is to reduce misdetections due to noise or clutter in the descriptor. If an extracted hash has a polluted bit and thus points to a wrong bucket, we might not retrieve the best view but still could recover from a similar view that we stored in the bucket the polluted hash points to. This strategy improves the robustness of the TBS retrieval, resulting in a consistently higher recall. We optimize the previous criterion with an additional term:
\begin{equation}
\argmin_B \frac{1}{|N_i|} \sum_i \Big| |\mathcal{S}_L^B(N_i) | - |\mathcal{S}_R^B(N_i)|  \Big| + 
\frac{1}{|N_i|^2} \Big( P(\mathcal{S}_L^B(N_i)) +
P(\mathcal{S}_R^B(N_i)) \Big) \;\;,
\end{equation}
where the second term  penalizes close views falling into the same side of the
split. We define the penalty function $P(\cdot)$  as:
\begin{equation}
P(N) := \sum_{\bx \in N} \sum_{\by \in N} \mathbb{I}(\bx,\by) \cdot
\begin{cases}
1,  \text{ if } \cos^{-1}(|\langle q_\bx,q_\by \rangle|)    < \tau \\
0,   \text{ otherwise} \;\; ,
\end{cases}
\end{equation}
where $\mathbb{I}(\bx,\by)$ indicates if descriptors $\bx$ and
$\by$ encode views of the same object and $q_\bx,q_\by$ are the quaternions associated with the rotational part of the descriptors' poses. We
set the proximity threshold $\tau=0.3$ empirically according to our viewpoint
sampling.

\subsection{Remarks on the Implementation}

For selecting the hash keys, we  rely on the descriptors after `bit spreading'
of  LineMOD~\cite{Hinterstoisser2012}, which  makes  the  descriptors robust  to
small  translations  and deformations  in  a  neighborhood  of $T$  pixels.   It
increases the  spatial overlap  of quantized  features and  allows for  a better
descriptor separability. For matching itself we used the unspreaded templates.

Furthermore, after one bit has been selected, we disallowed all bits closer than
$T$ to be selected for the same LineMOD value.  This forces the bit selection to
take different  values and positions into  account, as we sometimes  observed an
accumulation of  selected bits encoding the  same orientation in one  area which
could lead to bad recognition rates.

For efficiency,  we conduct the matching  analogously to \cite{Rios-Cabrera2013}
on  a fixed  grid.  As  opposed to  \cite{Hinterstoisser2012}, we  do not  use a
robust  cosine-based  similarity score  but  count  the  bits after  ANDing  the
descriptors  and  dividing by  the  number  of grid  points  falling  on the  view
foreground.

\section{Evaluation}
\label{sec:eval}

We ran our method on  the LineMOD ACCV12 dataset~\cite{Hinterstoisser2012} consisting of
15  objects and  followed  the  exact same  protocol  to  create an  equidistant
viewpoint sampling. Furthermore,  scale and  in-plane rotations
were sampled  accordingly to  cover a  pre-defined 6D  pose space,  resulting in
exactly $N=3115$ views per object. 


We  followed \cite{Hinterstoisser2012}  and spread  the quantized  values in  a
small neighborhood  of $T=8$  pixels which makes  the representation  robust and
allows to check only every $T$-th  image position.  Furthermore, we use the same
post-processing: after retrieval/matching, we sort  the candidates  according to
their score and run a rough color check to discard obvious outliers.  We conduct
a  fast  voxel-based  ICP~\cite{Fitzgibbon2001}  and reject  candidates  if  the
average euclidean  error is too large.  Finally, the first $n=10$  survivors are
projected onto  the scene  to run  a finer ICP  together with  a depth  check to
decide for  the best match.  We use the  same evaluation criteria with a distance
factor of $k_m=0.1$ to decide for a hit or miss.

The computational  cost during  testing is  modest: the whole  system runs  on a single  CPU-core---apart   from  the  post-processing  where   the  depth  check projections use OpenGL  calls---, uses no pyramid scheme and  the hash tables take up less than 1~MB.

\begin{figure*}
	\begin{center}
		\includegraphics[width=6cm]{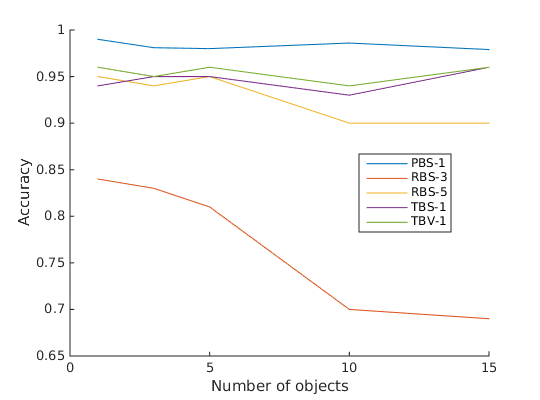}
		\includegraphics[width=6cm]{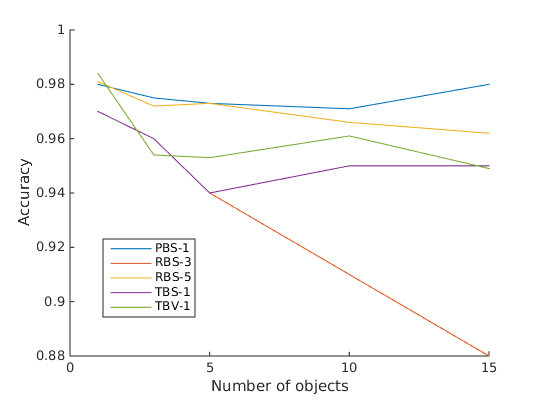}
		\includegraphics[width=6cm]{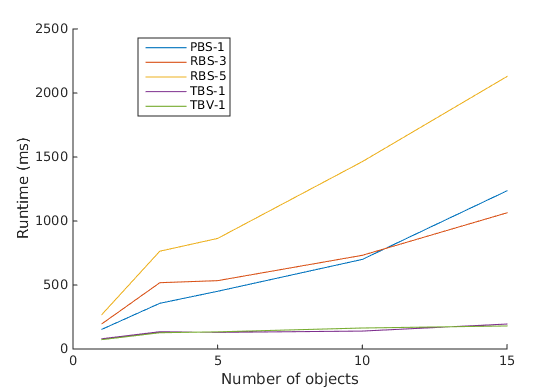}
		\includegraphics[width=6cm]{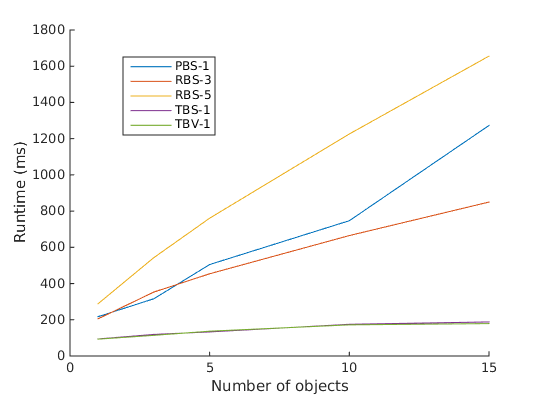}
	\end{center}
	\caption{Examples of  accuracies and runtimes  for different strategies  and a
		varying database  size. Left column:  for the `ape' sequence.  Right column:
		for the `lamp' sequence. The number  in the legend for each strategy denotes
		the amount of hash keys per window.}
	\label{fig:strats}
\end{figure*}

\renewcommand{\arraystretch}{1.2}

\begin{table}
	\footnotesize
	\begin{center}
		\begin{tabular}{@{}cccccc@{}}
			\toprule
			Seq / Objs & 1  & 3 & 5 & 10 & 15 \\ \midrule
			ape 		& 96.1\% / 73ms  & 95.1\% / 127ms & 96.8\% / 134ms & 94.1\% / 164ms & 95.6\% / 180ms \\
			bvise 		& 92.8\% / 83ms  & 95.4\% / 117ms & 91.2\% / 128ms&  92.3\% / 181ms &  91.2\% / 192ms\\ 
			bowl		& 99.3\% / 84ms  & 98.8\% / 114ms & 98.6\% / 123ms & 98.1\% / 156ms & 98.6\% / 184ms \\  
			cam			& 97.8\% / 75ms  & 96.9\% / 111ms & 95.3\% / 129ms & 94.8\% / 162ms & 95.2\% / 174ms \\  
			can 		& 92.8\% / 81ms  & 91.6\% / 112ms  & 93.7\% / 119ms  & 93.9\% / 158ms & 91.8\% / 171ms \\  
			cat 		& 98.9\% / 99ms  &  97.2\% / 117ms & 97.4\% / 138ms & 95.8\% / 164ms & 96.1\% / 188ms \\  
			cup 		& 96.2\% / 65ms  & 96.0\% / 100ms & 95.3\% / 117ms & 94.5\% / 144ms &  98.6\% / 194ms \\  
			driller 	& 98.2\% / 106ms  & 97.8\% / 135ms & 98.1\% / 162ms & 97.6\% / 171ms & 95.1\% / 190ms \\ 
			duck 		& 94.1\% / 74ms  & 90.8\% / 122ms & 90.7\% / 124ms & 91.5\% / 161ms & 92.9\% / 179ms \\  
			eggbox 	  & 99.9\% / 68ms  & 99.9\% / 103ms  & 99.9\% / 115ms & 99.9\% / 151ms & 99.9\% / 174ms \\  
			glue		  & 96.8\% / 100ms & 96.2\% / 131ms & 94.4\% / 154ms & 95.3\% / 166ms & 95.4\% / 175ms \\  
			hpuncher & 95.7\% / 97ms    & 95.3\% / 118ms & 95.8\% / 142ms & 95.2\% / 162ms & 95.9\% / 183ms \\  
			iron 		 & 96.5\% / 101ms  & 94.8\% / 122ms & 95.0\% / 141ms & 95.5\% / 167ms & 94.3\% / 203ms\\ 
			lamp 		& 98.4\% / 93ms    & 95.4\% / 114ms & 95.3\% / 137ms & 96.1\% / 172ms & 94.9\% / 179ms \\  
			phone 		& 93.3\% / 90ms   & 94.6\% / 126ms & 94.5\% / 133ms & 91.7\% / 167ms & 91.3\% / 198ms \\  
			\midrule
			TBV Average	& 96.5\% / 83ms  & 95.7\% / 117ms & 95.5\% / 131ms & 94.9\% / 162ms & 95.1\% / 184ms \\
			DTT-3D \cite{Rios-Cabrera2013}		& 97.2\% / 55ms  & 97.2\% / 81ms & 97.2\% / 107ms & 97.2\% / 173ms & 97.2\% / 239ms \\
			LineMOD \cite{Hinterstoisser2012}	& 96.6\% / 119ms & 96.6\% / 273ms & 96.6\% / 427ms & 96.6\% / 812ms & 96.6\% / 1197ms \\
			\bottomrule
		\end{tabular}
	\end{center}
	\caption{Accuracy and  runtime per frame for our whole pipeline with  the TBV
		strategy, DTT-3D and LineMOD for the whole dataset with a varying number of trained and loaded objects.  With only a few objects, DTT-3D  is  faster  than  our approach.  However,  its
		complexity increases linearly with the database size, allowing us to overtake when the number of objects becomes higher. Note that the dataset only provides groundtruth for one object per frame. }
	\label{table:acc_time_tbv}
\end{table}

\paragraph{Different learning strategies.}
We learned the  hashing for each sequence/objects configuration  by grouping the
object   of  interest   together  with   a  random   subset  of   the  remaining
ones. An exception  is the case for 15 objects  where we built the
hash tables  once and used  them for  all tests.  A  summary of our  evaluation is
given in Figure~\ref{fig:strats}. Note that  we conducted our experiments with a
varying amount of hash tables per window/scale  for each strategy but only plot the most
insightful to not  clutter the graphs and save space.  The behavior  was similar  across the whole dataset and we  thus present results for  all strategies only on  two sequences and then
restrict ourselves to the best strategy thereafter for a more detailed analysis.

The RBS  strategy was clearly  the weakest one.  This  is because RBS  managed a
poor separation  of descriptors: since the key bits were chosen  randomly, most
descriptors were  assigned a  hash value of pure zeros and were put into  the first
bucket while the rest of the hash  table was nearly empty.  This resulted during
testing in either  hitting an arbitrary bucket with no  elements or the 0-bucket
with a  high amount of  retrieved views,  approximating an exhaustive  search at
that image  location which increased matching  time.  It only started  to detect
accurately with multiple tables per window at the expense of very high runtimes.

Not  surprisingly,   PBS  nearly   always  managed   to  correctly   detect  the
object---limited  only  by  our  matching  threshold---while  achieving  similar
runtimes as RBS with  3 tables per window. An inspection  of the hashes revealed
that PBS led to multiple large buckets where descriptors concentrated and if one
of those  buckets was  hit, it  was very  likely that  it contained  the correct
view. Nonetheless, PBS does not take  advantage of all available buckets as some
of  them remain  empty and  therefore still  exhibits a  linear runtime  growth.
Using more tables for RBS was just slightly increasing runtime and accuracy.

For both TBV  and TBS the most  interesting observation is their  sublinear growth in
runtime.  Enforcing the  tables to be filled equally results  in an obvious drop
in the  amount of  retrieved views. Nonetheless,  both strategies  yield already
good accuracies  with one table per  window and  TBV was  able to  outperform TBS
usually with around $2\%$ in accuracy  since otherwise missed views could be
retrieved  and ICP-refined  to the  same  correct pose  from a  similar view  in
another bucket.

\paragraph{Comparison to related methods.} 
Since  TBV supplies  us  with a  sublinear runtime  growth  and acceptably  high
accuracies,  we settle  for  this strategy  and show  more  detailed results  in
Table~\ref{table:acc_time_tbv}. We are  able to  consistently
detect at around  $95\%-96\%$ accuracy on average which is slightly worse  than LineMOD and
DTT-3D. However, we are always faster than LineMOD and overtake DTT-3D at around
8 objects  where our constant-time  hashing overhead becomes negligible  and the
methods'  time complexities  dominate. This  is important  to stress  since real
scalability comes  from a sublinear  growth. Additionally, we show  more clearly
the scalability  of our approach when  increasing the amount of  descriptors: since
the dataset consists of only 15 objects, yielding 46,725 descriptors, we created
further 46,725 descriptors by drawing  each bit from its estimated distribution,
thus enlarging our database  artificially to 30 objects. Figure~\ref{fig:teaser}
shows  a graph  of our  runtimes in  comparison to  DTT-3D. For  the latter,  we
extrapolated  the  values given the  authors'
timings.  The gap  in  runtime  shows our  superiority  when  dealing with  many
objects and views.


\paragraph{Sublinear retrieval and matching.}

\begin{figure}
	\begin{center}
		\includegraphics[width=6cm]{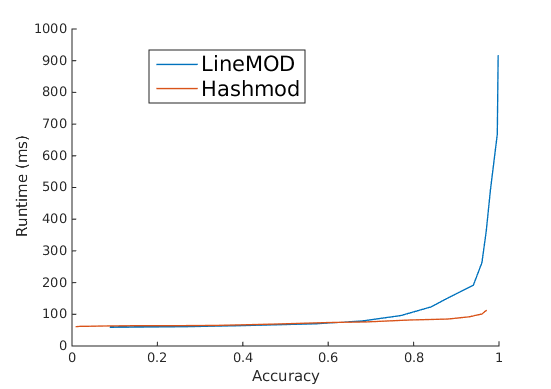}                   
		\includegraphics[width=6.5cm]{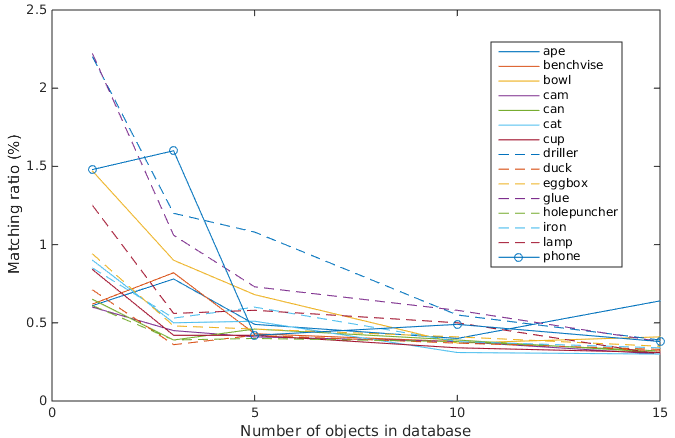}
	\end{center}
	\caption{Left: Accuracy versus runtime on the `driller'-sequence with TBV hash
		keys and LineMOD with a set of decreasing matching thresholds.
		Both methods  achieve  higher  accuracies with a lower threshold although we only  retrieve a fraction of views, making
		our runtime  increase marginal.   Right: Matching  ratios for  an increasing
		number of objects  using TBV hash keys. The obvious  decreasing trend allows
		us to scale with the number of objects.}
	\label{fig:ratios}
\end{figure}

After retrieval, we conduct template  matching together with an object-dependent
threshold.  Although this parameter is of  importance to balance runtime versus accuracy, we are less prone to
ill settings in comparison to LineMOD since  at each position we only retrieve a
tiny subset of candidates, as  shown in Figure~\ref{fig:ratios} (left). Obviously, the small set of retrieved views most often contains the correct one, leading to good accuracies while keeping the runtime low. Furthermore,  the ratio of total conducted matchings on an image of size $W \times H$,
\begin{equation}
\# \text{retrieved templates} \hspace{0.1cm} / \hspace{0.1cm} \frac{ \# \text{templates in database} \cdot W \cdot H}{T \cdot T}
\end{equation}
stays small as shown in  Figure ~\ref{fig:ratios} (right) and explains our improvement in
comparison to an exhaustive search: while increasing the object database size, the ratio grows smaller. It is this trend of decay that allows us to
scale sub-linearly with the number of objects/views.

\section{Conclusion and Acknowledgment}
We presented  a novel method for  3D object detection and  pose estimation which
employs hashing for efficient and  truly scalable view-based matching against RGB-D frames.  We
showed that we outperform the state-of-the-art in terms of speed while being able to achieve
comparable accuracies on a challenging dataset. It would be interesting to invest further effort into alternative hash learning schemes, different feature representations and extend the experiments to other challenging multi-object datasets.
The authors would also like to thank Toyota Motor Corporation for supporting
and funding the work.

\bibliography{short,cleaned_egbib}

\clearpage

\part*{Supplementary material}

\section{Different learning strategies}

As mentioned in the paper, the performance of the hash tables is governed by the length of the key, i.e. the number of buckets that are spanned, and by the distribution of the descriptors in the buckets. In Figure 7 we had a key length of $b=7$ which provided us with a table of 128 possible buckets. To relate to the discussed results in the evaluation, we present here the resulting tables acquired by each strategy. Furthermore, we provide 3 important quantities for each table: the actual number of used buckets, the largest number of descriptors that is in one bucket as well as the standard deviation computed over all non-empty buckets. 

Note that we did not necessarily take the tables from the same scale, since both TBV and TBS terminate prematurely if one leaf becomes empty after splitting, reducing the key length. We rather took the tables from different scales such that we can present each table spanning the same number of buckets to give a nicer qualitative comparison. Nonetheless, the behavior is the same across all objects and scales for each strategy, making our comparison unproblematic.

Apparently, RBS produced by far the largest disparity in bucket sizes where the largest, storing 76 descriptors, is also at the same time the mentioned 0-bucket. The remaining used 12 buckets are rather small and the other 115 are completely empty, leading to a high standard deviation. Hitting the 0-bucket retrieves a lot of descriptors, shooting up runtime unnecessarily while otherwise the method retrieves virtually nothing at all when hitting any other bucket.

PBS already uses more buckets than RBS and creates a more balanced separation into multiple larger sets. The best results are however obtained with our tree-based selection strategies that provide us with a good distribution among the buckets, i.e. small bucket sizes together with a small standard deviation. In the optimal case, the standard deviation would be zero and the buckets' usage would equal an uniform distribution.

\begin{figure*}[b]
	\centering
	\includegraphics[width=11cm]{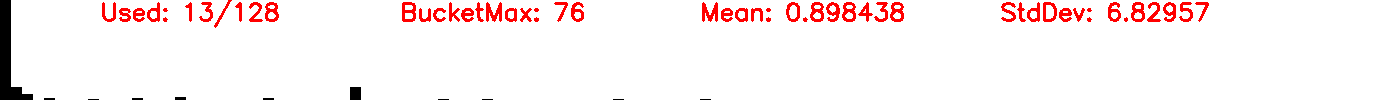}
	\hrule
	\vspace{0.5cm}
	\includegraphics[width=11cm]{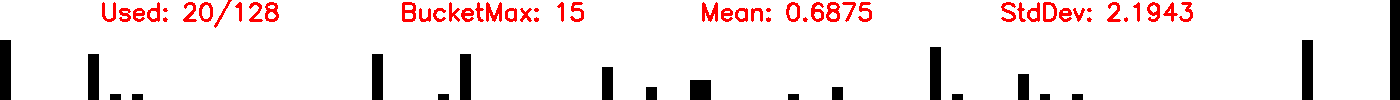}
	\hrule
	\vspace{0.5cm}
	\includegraphics[width=11cm]{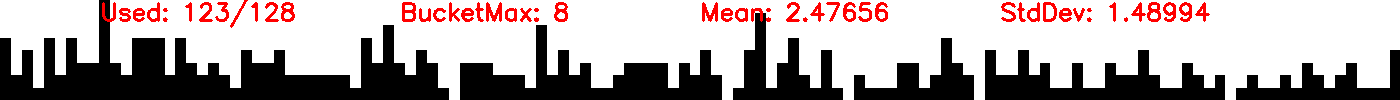}
	\hrule
	\vspace{0.5cm}
	\includegraphics[width=11cm]{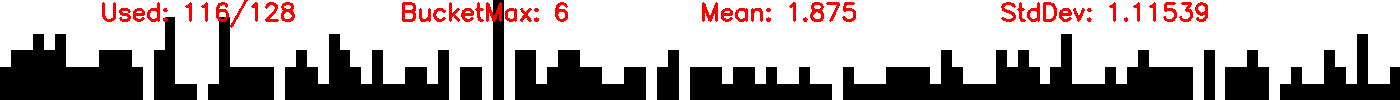}
	\hrule
	\vspace{0.1cm}
	\label{fig: strats}
	\caption{For each strategy, we depict a filled hash table that we computed for the `benchvise'. From top to bottom: RBS, PBS, TBS, TBV. Note that the height for each hash table is normalized, skewing the comparison optically.}
\end{figure*}

\section{Matching thresholds}
In order to determine each object-dependent matching threshold for our grid-based similarity score, we conducted a sweep over the thresholds for each sequence and fixed them such that each object would be found at least with an average accuracy of 98\% in an exhaustive search scenario. See Figure 8 for the graphs.

\begin{figure*}
	\centering
	\includegraphics[width=11cm]{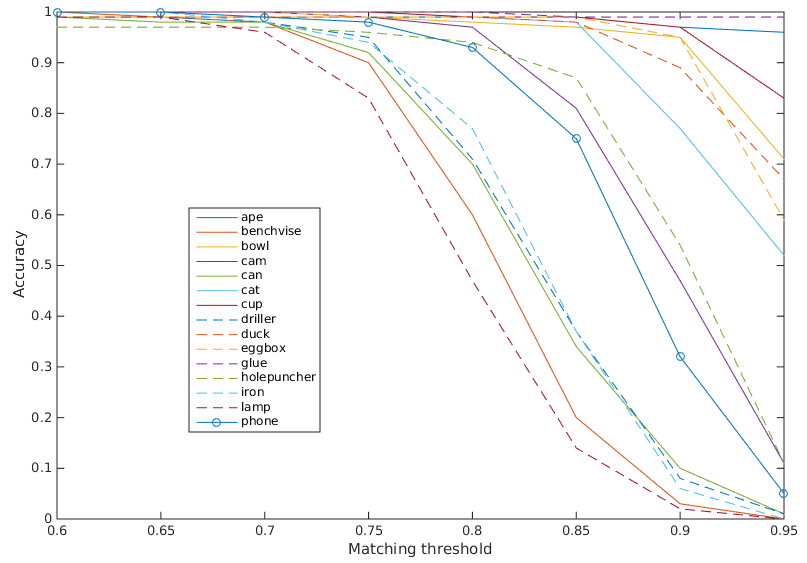}
	\label{fig: exh}
	\caption{Average accuracy for each object on its associated sequence with different matching thresholds when running an exhaustive search.}
\end{figure*}

\section{Different number of tables per scale}

We evaluated each strategy with a different amount of trained hash tables per scale in Figure 9. For this, the descriptors for each scale were subdivided into multiple overlapping groups and then one table was responsible for indexing into one of them. As we can observe on the next page, increasing the amount of tables always leads to better recognition at the expense of higher runtimes. Except for the RBS strategy which would have needed even more tables still, all methods experienced a saturation in detection accuracy for the `ape' around $99.5\% $. In light of to these results, we settled for TBV and one table per scale because it allowed us to be faster than DTT-3D with a comparable accuracy.
\begin{figure*}
	\centering
	\includegraphics[width=6cm]{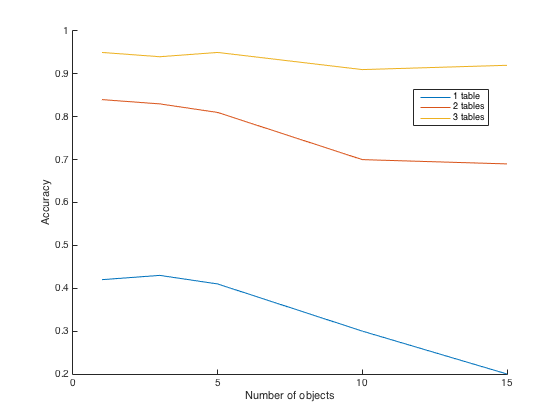}
	\includegraphics[width=6cm]{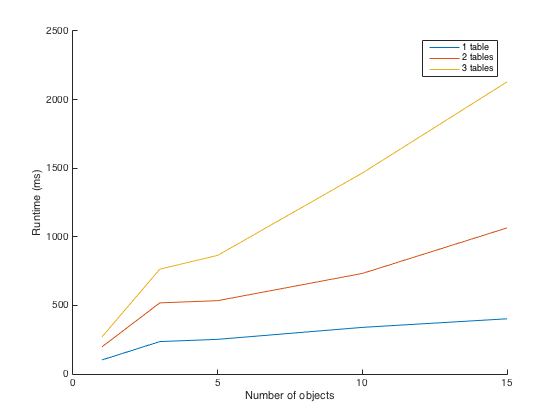}
	\includegraphics[width=6cm]{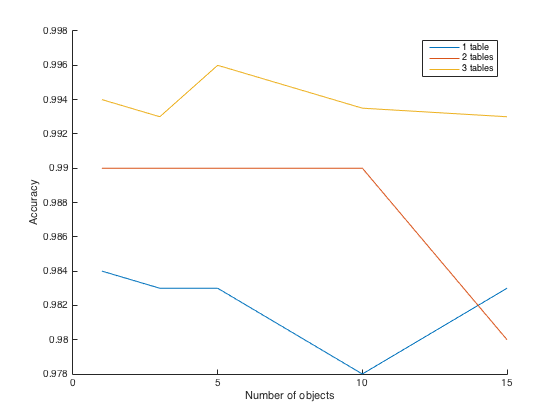}
	\includegraphics[width=6cm]{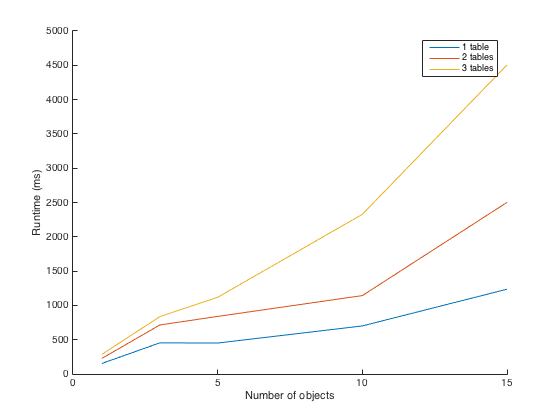}
	\includegraphics[width=6cm]{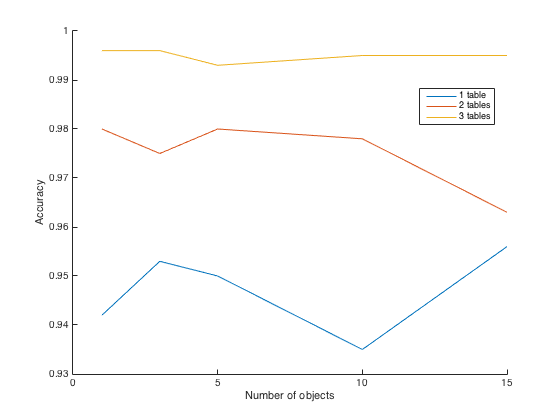}
	\includegraphics[width=6cm]{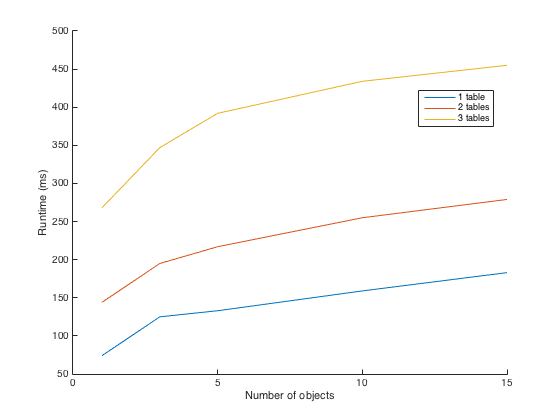}
	\includegraphics[width=6cm]{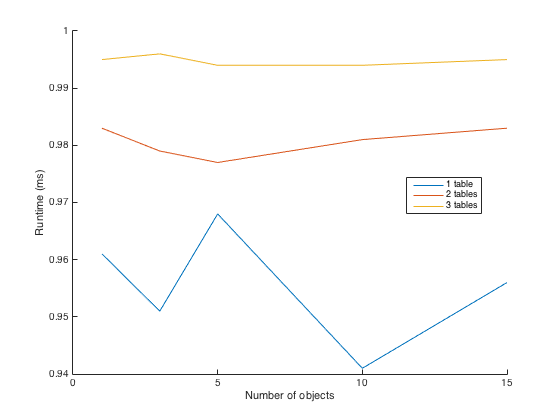}
	\includegraphics[width=6cm]{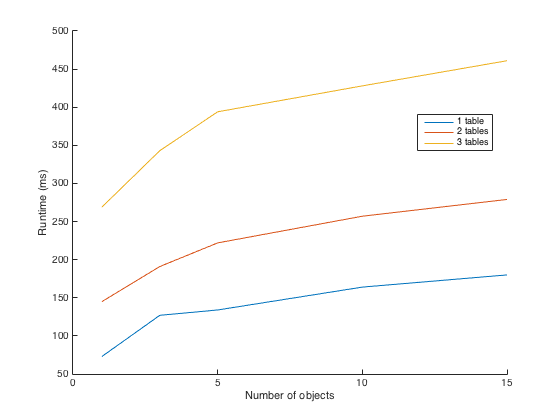}
	\label{fig: large}
	\caption{Plotting the behavior for each strategy on the `ape' sequence when the amount of tables per scale increases. From top to bottom: RBS, PBS, TBS, TBV.}
\end{figure*}

\end{document}